\documentclass[pdflatex,sn-vancouver,Numbered]{sn-jnl}

\usepackage{graphicx}%
\usepackage{multirow}%
\usepackage{amsmath,amssymb,amsfonts}%
\usepackage{amsthm}%
\usepackage{mathrsfs}%
\usepackage[title]{appendix}%
\usepackage{xcolor}%
\usepackage{textcomp}%
\usepackage{manyfoot}%
\usepackage{booktabs}%
\usepackage{algorithm}%
\usepackage{algorithmicx}%
\usepackage{algpseudocode}%
\usepackage{listings}%

\begin{document}

\title[Article Title]{Estimating the stability number of a random graph using convolutional neural networks}

\author*[1,2]{\fnm{Randy} \sur{Davila}}\email{rrd6@rice.com}


\affil*[1]{\orgdiv{Department of Computational Applied Mathematics \& Operations Research}, \orgname{Rice University}, \orgaddress{\street{6100 Main Street}, \city{Houston}, \postcode{77005}, \state{Texas}, \country{USA}}}

\affil[2]{\orgdiv{Research and Development}, \orgname{RelationalAI}, \orgaddress{\street{ 2120 University Avenue}, \city{Berkeley}, \postcode{94704}, \state{California}, \country{USA}}}



\abstract{Graph combinatorial optimization problems are widely applicable and notoriously difficult to compute; for example, consider the traveling salesman or facility location problems. In this paper, we explore the feasibility of using convolutional neural networks (CNNs) on graph images to predict the cardinality of combinatorial properties of random graphs and networks. Specifically, we use image representations of modified adjacency matrices of random graphs as training samples for a CNN model to predict the stability number of random graphs; where the stability number is the cardinality of a maximum set of vertices in a graph that contains no pairwise adjacency between vertices. The model and results presented in this study suggest potential for applying deep learning in combinatorial optimization problems previously not considered by simple deep learning techniques.}

\keywords{Convolutional neural networks, independence number, stable sets, stability number.}



\maketitle

\section{Introduction}\label{sec1}
Combinatorial optimization (CO) is a heavily studied and widely applicable subfield of optimization that combines techniques in combinatorics, linear programming, and the theory of algorithms to solve discrete optimization problems; see, for example, the excellent text by Cook, Cunningham, Pulleyblank, and Schrijver~\cite{CO}. Some of the most well-known problems in this field include the knapsack problem, the traveling salesman problem, the graph coloring problem, the matching problem, and the facility location problem. In this paper, we consider another well-known CO problem called \emph{maximum stable set problem}, and more specifically, we provide evidence for using deep learning techniques in its solution. The maximum stable set problem concerns computing the largest set of independent vertices in a graph and is a well-known NP-complete problem~\cite{stable-NP}. The associated \emph{stability number} of a graph $G$, denoted $\alpha(G)$, is the cardinality of a maximum stable set in $G$. Calculating this parameter is computationally expensive for large graphs, as it belongs to the class of NP-hard computable graph parameters~\cite{stable-NP}.

We propose a novel method to accurately estimate the stability number of random graphs using feature engineering and \emph{convolutional neural networks} (CNNs). This approach combines the strengths of CNNs in image recognition with graph theoretical concepts to provide a new perspective on graph parameter estimation. Our main contributions in this paper are as follows:
\begin{itemize}
    \item[(1)] Introduce a graph feature engineered representation suitable for training CNNs.
    \item[(2)] Provide a CNN model that approximates the stability number for random graphs. 
    \item[(3)] Provide evaluation metrics on our trained model demonstrating viability for CNNs in the solution of the maximum stable set problem.
\end{itemize}

The rest of this paper is organized as follows. Section~\ref{sec:related-work} provides graph terminology and background on the maximum stable set problem, as well as a brief discussion on CNNs. Section~\ref{sec:methodology} describes training data generation, our feature engineering approach, computation of the stability number of a graph, CNN model architecture, and code used for these processes. Section~\ref{sec:results} gives evaluation metrics returned by our trained model. Section~\ref{sec:discussion} concludes this paper with a short discussion of the work presented in this paper. 

\section{Related Work}\label{sec:related-work}
Throughout this paper, all graphs will be considered simple, undirected, and without loop edges. Specifically, we let $G = (V, E)$ be a graph with vertex set $V$ and edge set $E$. A set $S\subseteq V$ of vertices is called an \emph{stable set} (also commonly called an \emph{independent set}) if no two vertices in $S$ form an edge in $E$. The cardinality of a maximum stable set in $G$ is called the \emph{stability number} (also commonly called the \emph{independence number}) of $G$ and is denoted $\alpha(G)$. The stability number is one of the oldest and most heavily studied parameters in graph theory; for example, see~\cite{stable-chapter} in the excellent graph theory textbook by Bondy and Murty~\cite{GT}. Theoretically, the stability number is related to various set coverings in graphs and the largest cliques. From a practical standpoint, detecting maximum stable sets has many applications, such as in case-based reasoning systems~\cite{reason}, computer vision~\cite{vision}, molecular biology~\cite{bio1, bio2, bio3, bio4}, and in transmitting messages~\cite{GT}. Furthermore, the computation of the stability number for a graph is known to be NP-hard (see~\cite{stable-NP}), and so, much work has been put into approximating this number for a given graph. 

The automatic analysis of image data is the focus of computer vision and represents one of the significant areas in machine learning. Convolutional neural networks (CNNs) have demonstrated the robust ability to interpret the structured information contained in image data accurately; see, for example, the comprehensive text on neural networks by Bishop and Bishop~\cite{bishop}. Unlike simple \emph{multilayer perceptrons} (MLPs), CNNs make use of the spatial 2-dimensional structure of images by way of feature detectors -- \emph{convolutional filters} and \emph{pooling operations}. As a consequence, CNNs are translational invariant, meaning that shifting and rotating an image will not change the interpretation of the trained CNN model. Recently, there have been many advancements in the study of deep neural networks applied to graph-structured data~\cite{GNN-review}, a field that generally falls under the umbrella of \emph{graph neural networks} (GNNs). \emph{Graph Convolutional Networks} (GCNs) are a variant of GNNs and CNNs explicitly designed for analysis on graph-structured data that make use of the convolution operation becoming the \emph{de facto} method for implementing neural networks on graphs~\cite{pmlr-v97-wu19e}.

To our knowledge, no published studies have attempted to approximate the stability number of a graph using only images of the graph. Moreover, our approach differs from those attempted by GNNs and GCNs as we interpret graphs as images instead of attempting to learn the representations directly from the graph relational structure with a neural network. Notably, our technique makes use of feature engineering, a common practice before the success of neural networks. We demonstrate in the following sections that this technique seems suitable for some combinatorial optimization problems on graphs by way of studying CNNs applied to computing a graph's stability number. 

\section{Methodology}\label{sec:methodology}
\begin{figure}[htbp]
    \centering
    \includegraphics[width=\linewidth]
    {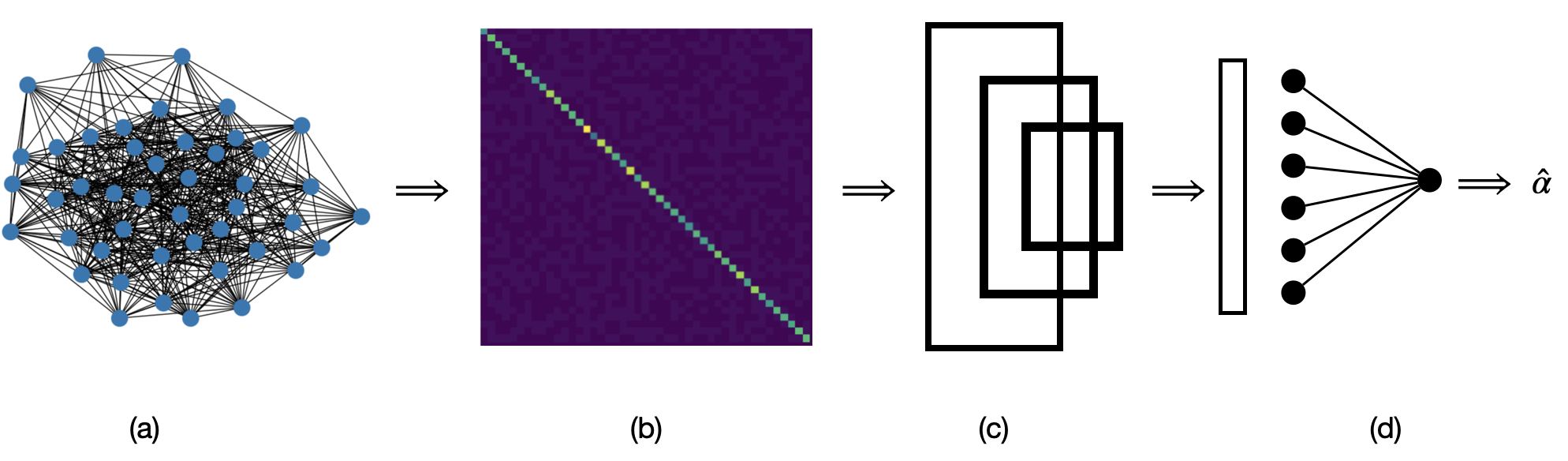}
    \caption{Our model for inputing a random graph into a CNN and predicting the stability number. (a) a random graph is presented. (b) a modified adjacency matrix image is constructed. (c) the image representation of the graph is filtered using convolutional layers in a CNN. (d) the resulting data produced from the convolutional layers is flattened and feed into a shallow MLP producing a numerical value $\hat{\alpha}$. }
    \label{fig:model-flow}
\end{figure}

Our hypothesis is that a visual representation of a simple graph will allow a CNN to train sufficiently in order to predict the stability number of a graph. The steps for producing a predicted value for the stability number of a graph that we implement are shown in Fig~\ref{fig:model-flow}, where a graph is first converted into an image, which is then fed into a CNN model. 
The computations and experiments performed are implemented in Python and are readily available at this \href{https://github.com/RandyRDavila/Estimating-the-stability-number-of-a-random-graph-using-convolutional-neural-networks}{GitHub repository}.

\subsection{Data Generation}
To train our CNN, we first need to generate a dataset of graphs with known stability numbers. This process involves several steps to ensure a diverse and representative dataset for effective model training. Namely, the generation of random graphs, the mapping of each graph to our image representation, and finally labeling of each graph with their respective stability number.

The random graphs that we choose to use for our experiments are generated using the \emph{Erdős–Rényi model} and notation $G(n, p)$, where $n$ is the number of vertices and $p$ is the probability of edge formation between any two vertices. The Erdős–Rényi model is a well-known method for generating random graphs and provides a straightforward mechanism for creating graphs with varying densities \cite{erdos-renyi}. In our experiments, we make use of the NetworkX library in Python to generate these Erdős–Rényi random graphs~\cite{networkx}. For our dataset, we vary the number of vertices $n$ up to a maximum of 64 and use different values of $p$ to create graphs with different edge densities. This variation ensures that the training set includes a wide range of graph structures, while also leaving many unseen graph structures for testing. Specifically for training and testing our CNN model, we generated a dataset of 2,000 random graphs with node sizes ranging from 10 to 64 and with edge probability chosen randomly. The dataset was split into training and testing sets, with 80\% used for training and 20\% for testing.

To label our training and testing data we make use of an optimization technique for computing the stability number of a graph precisely. Even though computation of $\alpha(G)$ is intractable in general, we may still compute it exactly for graphs with small order by providing the following linear-integer optimization formulation to open source or commercial solvers:

\textbf{Maximum Stable Set Linear-Integer Program Formulation.} Given a graph $G = (V, E)$, we may find an optimal (maximum) stable set of $G$ by solving the following linear-integer program.
\begin{align}
\text{maximize} \quad & \sum_{i \in V} x_i \\
\text{subject to} \quad & x_i + x_j \leq 1 \quad \forall \, (i, j) \in E \\
                        & x_i \in \{0, 1\} \quad \forall \, i \in V
\end{align}
where, $x_i$ is a binary variable that indicates whether vertex $i$ is included in the independent set. The objective function $\sum_{i \in V} x_i$ aims to maximize the number of vertices in the independent set. The constraint $x_i + x_j \leq 1$ ensures that no two adjacent vertices are included in the independent set, and $x_i \in \{0, 1\}$ enforces that the variables are binary. Note, this optimization model is only feasible for relatively small order as the complexity of solving linear-integer programs is NP-hard.

\begin{figure}[htbp]
    \centering
    \includegraphics[width=\linewidth]
    {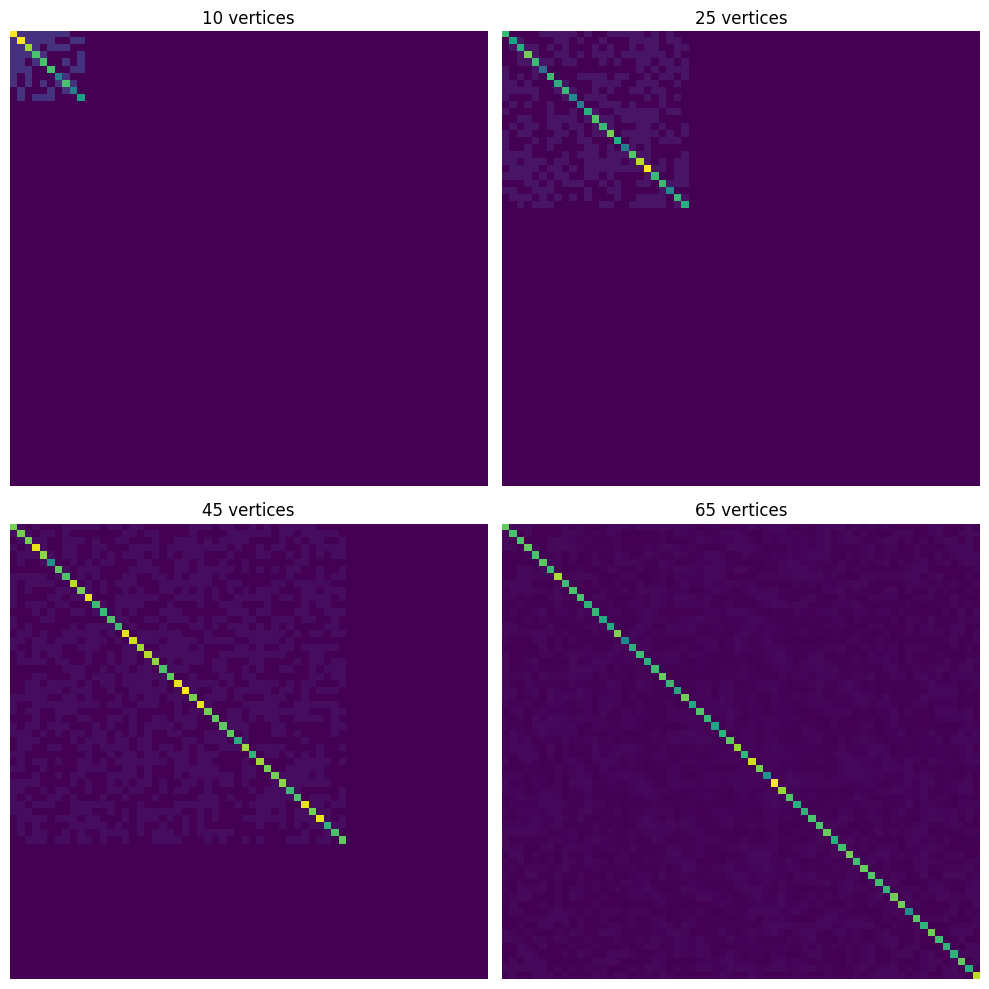}
    \caption{Image representations of random graphs with varying orders $n \in \{ 10, 25, 45, 65\}$. Note the dark shaded regions of the padded images for graphs with order $n < 65$. }
    \label{fig:graph-images}
\end{figure}

Once the graphs are generated and labeled with their respective stability numbers, we next construct their adjacency matrices. Recall that the adjacency matrix, namely the matrix $A$, such that $A_{i, j} = 1$ when vertex $i$ is adjacent to vertex $j$, and zero otherwise. This matrix can easily be converted to an image using matplotlib~\cite{Hunter:2007}, and we do so, but also choose additional steps in our mapping of the graph to an image suitable for training a CNN for computing the stability number of a graph. First, to maintain consistency across different graphs of different orders, each adjacency matrix is resized to a fixed dimension of $64 \times 64$ pixels. If the matrix is smaller than the target size, it is padded with zeros to reach the desired dimensions. Conversely, larger matrices are resized using \emph{bilinear interpolation} to reduce their size. Second, we place on the diagonal of the adjacency matrix $A$, entries that map the degree of the associated vertex to a heat map for visualization; and so, provide an image that highlights the  degree of each vertex of the graph explicitly; see Algorithm~\ref{alg:convert_heatmap} for this procedure and see Fig.~\ref{fig:graph-images} for example images of a few graphs.

\begin{algorithm}
\caption{Convert Graph to Heatmap Image}
\label{alg:convert_heatmap}
\begin{algorithmic}[1]
\Require Graph $G = (V, E)$, target image size $64 \times 64$
\Ensure Heatmap image representation of the graph
\State Obtain the adjacency matrix $A$ of the graph $G$
\State Initialize a zero matrix $P$ of size $64 \times 64$
\If{size of $A$ is less than $64 \times 64$}
    \State Pad $A$ with zeros to create $P$
\ElsIf{size of $A$ is greater than $64 \times 64$}
    \State Resize $A$ using bilinear interpolation to create $P$
\Else
    \State $P \gets A$
\EndIf
\State Compute the degree matrix $D$ with degrees of nodes on the diagonal
\State Add $D$ to $P$ to highlight node degrees
\State Normalize the combined matrix $P$
\State Convert the normalized matrix to an image with pixel values scaled to $[0, 255]$
\State Return the resulting heatmap image
\end{algorithmic}
\end{algorithm}

\subsection{CNN Architecture}
Our CNN model is designed to process graph images like those shown in Fig~\ref{fig:graph-images}, and predict the stability number given by the linear-integer program formulation presented prior; essentially a regression task given a graph as input. Our CNN consists a simple architecture like that first introduced in the text by Collet~\cite{chollet2021deep} (using \emph{Tensorflow}). That is, our model consists of multiple convolutional layers that apply convolutional filters to extract features from the image representation of the graph. 

The first convolutional layer has 32 filters of size \(3 \times 3\), followed by a ReLU activation function. This is followed by a max pooling layer of size \(2 \times 2\). Two more convolutional layers with 64 filters each of size \(3 \times 3\) and ReLU activation functions are added, each followed by a max pooling layer of size \(2 \times 2\). The pooling layers are used to reduce the dimensionality of the feature maps, retaining the most important information. The pooling operation helps in reducing the spatial dimensions, thus minimizing the computational load and controlling overfitting. The output of the convolutional layers is flattened into a one-dimensional vector. This flattened vector serves as the input to the fully connected layers. The flattened vector is fed into a small MLP. We use a dense layer with 64 units and ReLU activation function, followed by an output layer with a single neuron to predict the stability number. The final layer does not use an activation function since this is a regression task.

\subsection{Training Parameters}
Our full set of data consists of 2,000 random graphs $G(n, p)$ with $5 \leq n \leq 64$ and $p$ chosen uniformly and randomly; for which 20\% we leave out for testing. The training of our model uses the Adam optimizer and mean squared error as the loss function. The model is compiled with these settings and trained on the training dataset for 15 epochs with a validation split of 20\%. The performance metric used during training is the mean absolute error (MAE): This metric measures the average of the errors. It is more sensitive to large errors and is given by
\[
\text{MSE} = \frac{1}{n} \sum_{i=1}^{n} | \alpha_i - \hat{\alpha}_i |,
\]
where $\alpha_i$ is the true stability number and $\hat{\alpha}_i$ is the predicted stability number for the $i$-th graph $G$ in the testing dataset.

\section{Experimental Results}\label{sec:results}
The process of building and then training our model for predictions is given by Algorithm~\ref{alg:cnn_stability} which is shown below. 
\begin{algorithm}
\caption{Predicting Stability Number Using CNNs}
\label{alg:cnn_stability}
\begin{algorithmic}[1]
\Require Random graph $G = (V, E)$
\Ensure Predictions for the stability number
\State Set batch size to 64, optimizer Adam (learning rate: $1 \times 10^{-3}$), epochs number $n$ to 15
\State Generate dataset of random graphs and calculate their stability numbers
\State Convert graphs to images
\State Split dataset into training and testing sets
\State Define CNN model with input shape $(64, 64, 1)$
\For{each epoch $i$ from 1 to $n$}
    \State Train the CNN model on the training set
\EndFor
\State Use the trained model to predict stability numbers for new random graphs
\end{algorithmic}
\end{algorithm}
The training and validation metrics for each epoch during the training of CNN are summarized in Table~\ref{tab:training_metrics}.
\begin{table}[h!]
\centering
\begin{tabular}{@{} lcccc @{}}
\toprule
\textbf{Epoch} & \textbf{Training Loss MSE} & \textbf{Training MAE} & \textbf{Validation Loss MSE} & \textbf{Validation MAE} \\
\midrule
1  & 90.5996 & 4.9867 & 33.2031 & 3.9745 \\
2  & 9.9340  & 2.0060 & 14.8803 & 1.6592 \\
3  & 4.5899  & 1.3247 & 7.4233  & 1.6219 \\
4  & 3.4002  & 1.1684 & 7.4586  & 1.2359 \\
5  & 3.0771  & 1.0926 & 7.6951  & 1.3172 \\
6  & 2.7531  & 1.0659 & 9.0520  & 1.3086 \\
7  & 3.0603  & 1.1429 & 6.3911  & 1.1527 \\
8  & 3.7142  & 1.2327 & 7.6028  & 1.2700 \\
9  & 1.7859  & 0.8874 & 6.3903  & 1.0982 \\
10 & 1.5239  & 0.8160 & 10.5261 & 1.4262 \\
11 & 1.2684  & 0.7495 & 7.1824  & 1.1218 \\
12 & 1.1535  & 0.7028 & 9.4655  & 1.1884 \\
13 & 1.1565  & 0.7252 & 8.8410  & 1.2789 \\
14 & 1.1576  & 0.7053 & 8.2104  & 1.2372 \\
15 & 1.6186  & 0.8522 & 10.3406 & 1.4398 \\
\bottomrule
\end{tabular}
\caption{Training and validation metrics for each epoch during the training of the original model.}
\label{tab:training_metrics}
\end{table}
The performance of our model for predicting the stability number of random graphs in the test set was evaluated using several metrics, including mean squared error (MSE), mean absolute error (MAE), root mean squared error (RMSE), and R-squared (R²). Additionally, a 95\% confidence interval was calculated for the predictions. The evalutation metrics are summarized in Table \ref{tab:results}.

\begin{table}[h!]
\centering
\begin{tabular}{@{} lcc @{}}
\toprule
\textbf{Metric} & \textbf{Value} \\
\midrule
Test MSE & 3.4011785132213004 \\
Test MAE & 1.1447576931118966 \\
Test RMSE & 1.8442284330367809 \\
Test R-squared & 0.9320189094195505 \\
95\% Confidence Interval for the Predictions & (8.001530940870595, 9.488640491624523) \\
\bottomrule
\end{tabular}
\caption{Summary of evaluation results for the trained CNN model.}
\label{tab:results}
\end{table}

In our model, the MSE on the test set is close to 3.4; a relatively low value indicating that the model's predictions are generally close to the actual stability numbers. Our model achieved a MAE of roughly 1.145, suggesting that, on average, the model's predictions are about 1.145 units away from the actual stability numbers. The RMSE value for our model is about 1.84. The model has an R² value of about 0.932, signifying that approximately 93.2\% of the variance in the stability numbers can be explained by the model. Our trained model also has a confidence interval is roughly (8.001, 9.49).

\subsection{Runtime Comparison to the Integer Linear Program Solution}
To compare the computational efficiency and accuracy of the CNN predictions with the ineteger linear programming solution, we conducted experiments on random graphs of varying sizes. We measured the time taken to compute the stability number using both methods and recorded the difference between the true stability number and the predicted number by the CNN. The average results are summarized in Table \ref{tab:comparison}.

\begin{table}[h!]
\centering
\begin{tabular}{@{} lcc @{}}
\toprule
\textbf{Metric} & \textbf{ILP Method} & \textbf{CNN Method} \\
\midrule
Average Time (seconds) & 2.8774 & 0.0626 \\
Average stability Number & 8.9000 & 8.3031 \\
\bottomrule
\end{tabular}
\caption{Comparison of average runtime and stability number between ILP and CNN methods over 10 trials.}
\label{tab:comparison}
\end{table}

The results show a significant difference in the computational efficiency between the ILP method and the CNN prediction. The ILP method takes an average of 2.8774 seconds to compute the stability number, whereas the CNN method takes only 0.0626 seconds on average. This demonstrates the potential of using CNNs for rapid estimation of combinatorial properties in graphs. However, the accuracy of the CNN prediction is slightly lower than the exact computation by the ILP method. The average stability number predicted by the CNN is 8.3031, compared to the exact average of 8.9000 computed by the ILP method. Despite this discrepancy, the CNN method provides a reasonably close estimate with a much faster computation time.

\begin{figure}[h!]
\centering
\includegraphics[width=\textwidth]{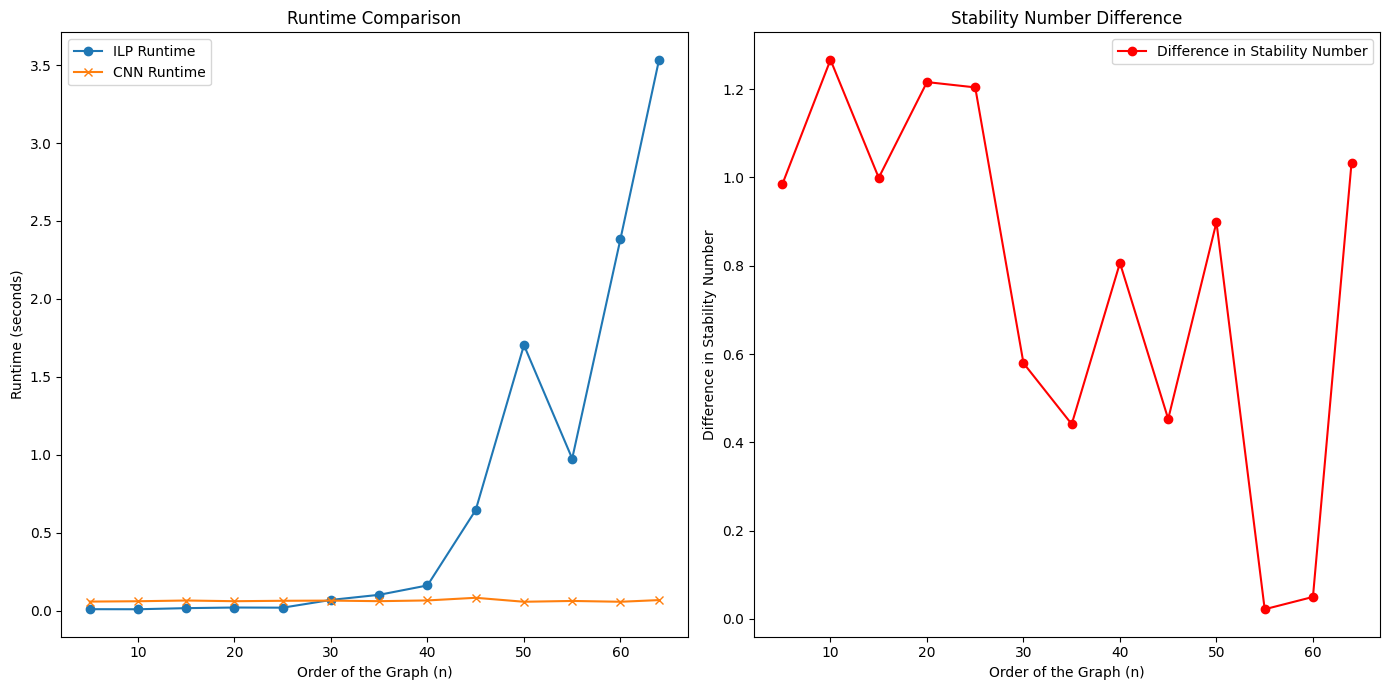}
\caption{(Left) Runtime comparison between ILP and CNN methods for varying graph sizes. (Right) Difference between the true stability number obtained by the ILP and the CNN predicted stability number -- notably the CNN predictions all are within a 1.0 of the true value.}
\label{fig:runtime_accuracy}
\end{figure}

To further illustrate the performance differences, we plotted the runtime and the difference in the stability numbers for varying graph sizes. Figure \ref{fig:runtime_accuracy} shows the runtime comparison and the stability number difference as the order of the graph increases.
The runtime comparison plot shows that the ILP method's runtime increases significantly with the order of the graph, while the CNN method remains relatively constant. The accuracy plot indicates that the difference in the stability number between the two methods is relatively small and consistent across different graph sizes.

The results show a significant difference in the computational efficiency between the ILP method and the CNN prediction. The ILP method takes an average of 2.8774 seconds to compute the stability number, whereas the CNN method takes only 0.0626 seconds on average. This demonstrates the potential of using CNNs for rapid estimation of combinatorial properties in graphs. However, the accuracy of the CNN prediction is slightly lower than the exact computation by the ILP method. The average stability number predicted by the CNN is 8.3031, compared to the exact average of 8.9000 computed by the ILP method. Despite this discrepancy, the CNN method provides a reasonably close estimate with a much faster computation time.

These findings highlight the trade-off between computational efficiency and accuracy when using neural networks for combinatorial optimization problems. The CNN method offers a viable solution for scenarios where rapid approximations are preferred over exact solutions, especially in large-scale applications or when many different graph stability numbers need to be computed.

\subsection{Ablation Study: Impact of Pooling Layers}
To assess the contribution of the pooling layers in our convolutional neural network (CNN) model, we conducted an ablation study where we removed the pooling layers and evaluated the model's performance. The training and validation loss (Mean Squared Error - MSE) and Mean Absolute Error (MAE) were recorded over 15 epochs, and the final test MAE was compared with the original model. The following table summarizes the training and validation loss and MAE for each epoch:

\begin{table}[h!]
\centering
\begin{tabular}{@{} lcccc @{}}
\toprule
\textbf{Epoch} & \textbf{Training Loss MSE} & \textbf{Training MAE} & \textbf{Validation Loss MSE} & \textbf{Validation MAE} \\
\midrule
1  & 3132.2964 & 16.5901 & 121.3369 & 8.2753 \\
2  & 116.5330  & 8.0357  & 121.2030 & 8.2672 \\
3  & 116.3825  & 8.0264  & 121.0281 & 8.2567 \\
4  & 116.1972  & 8.0150  & 120.8261 & 8.2444 \\
5  & 115.9861  & 8.0019  & 120.5956 & 8.2304 \\
6  & 115.7513  & 7.9870  & 120.3423 & 8.2150 \\
7  & 115.4974  & 7.9708  & 120.0667 & 8.1982 \\
8  & 115.2229  & 7.9539  & 119.7802 & 8.1807 \\
9  & 114.9353  & 7.9356  & 119.4721 & 8.1619 \\
10 & 114.6296  & 7.9167  & 119.1580 & 8.1426 \\
11 & 114.3142  & 7.8966  & 118.8228 & 8.1220 \\
12 & 113.9850  & 7.8756  & 118.4752 & 8.1006 \\
13 & 113.6442  & 7.8539  & 118.1193 & 8.0786 \\
14 & 113.2941  & 7.8316  & 117.7514 & 8.0558 \\
15 & 112.9336  & 7.8085  & 117.3775 & 8.0326 \\
\bottomrule
\end{tabular}
\caption{Training and validation performance over 15 epochs without pooling layers.}
\label{tab:ablation_results}
\end{table}

After training for 15 epochs, the model was evaluated on the test set. The test MAE for the model without pooling layers was 8.0051. The ablation study results indicate that the pooling layers significantly contribute to the model's performance. Specifically:

\begin{itemize}
    \item The training loss and MAE decreased steadily over the epochs, indicating that the model was learning even without the pooling layers.
    \item The validation loss and MAE followed a similar trend, but the final values remained high, suggesting that the model was not generalizing well to unseen data.
    \item The final test MAE (8.0051) is notably higher than the original model's test MAE (1.1448), demonstrating the importance of pooling layers in improving the model's accuracy.
\end{itemize}

\section{Discussion}\label{sec:discussion}
\begin{figure}[htbp]
    \centering
    \includegraphics[width=\linewidth]
    {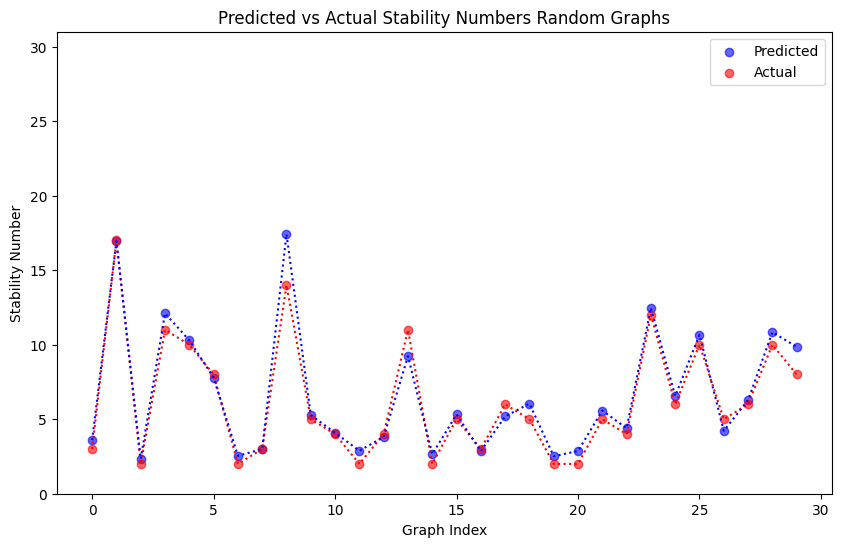}
    \caption{A sample of 30 random graphs $G(n, p)$, with order $n=30$ and edge probability $p$ chosen randomly. Predicted stability numbers shown in blue. True stabilty numbers shown in red. }
    \label{fig:predicted-vs-true}
\end{figure}

The low values of MSE, MAE, and RMSE combined with a high R² value indicate that our CNN model performs well in predicting the stability number of random graphs. The narrow confidence interval further supports the reliability of our model's predictions. These results demonstrate the effectiveness of using CNNs in the context of graph combinatorial optimization problems, specifically in estimating the stability number of random graphs. For example, see Fig.~\ref{fig:predicted-vs-true}, which demonstrates the predictive properties of our trained CNN on a sample of 30 random graphs $G(n, p)$, each with order $n=30$ and edge probability $p$ chosen randomly.

Our study also demonstrates how trained CNNs significantly reduce the computation time for the stability number when compared to traditional methods like integer programming. While the CNN's predictions are not exact, they are close enough to be useful in practical applications where speed is crucial. This approach opens up new possibilities for solving other challenging graph theory and CO problems using simple deep learning techniques.

\subsection*{Code and Data availability}
All code used in this paper can be found at this \href{https://github.com/RandyRDavila/Estimating-the-stability-number-of-a-random-graph-using-convolutional-neural-networks}{GitHub repository} under the notebooks directory.

\subsection*{Conflicts of interest}
The authors declare no competing interests.


\bibliography{sn-bibliography}

\end{document}